\documentclass[10pt,twocolumn,letterpaper]{article}

\usepackage{iccv}
\usepackage{times}
\usepackage{epsfig}
\usepackage{graphicx}
\usepackage{amsmath}
\usepackage{amssymb}
\usepackage{subcaption}

\usepackage[breaklinks=true,bookmarks=false]{hyperref}

\iccvfinalcopy 

\def\iccvPaperID{15} 

\begin{document}

\title{Render4Completion: Synthesizing Multi-view Depth Maps for 3D Shape Completion}

\author{Tao Hu, Zhizhong Han, Abhinav Shrivastava, Matthias Zwicker\\
	University of Maryland, College Park\\
	{\tt\small taohu@cs.umd.edu, h312h@umd.edu, abhinav@cs.umd.edu, zwicker@cs.umd.edu}
}

\maketitle

\begin{abstract}
	We propose a novel approach for 3D shape completion by synthesizing multi-view depth maps. While previous work for shape completion relies on volumetric representations, meshes, or point clouds, we propose to use multi-view depth maps from a set of fixed viewing angles as our shape representation. This allows us to be free of the memory limitations of volumetric representations and point clouds by casting shape completion into an image-to-image translation problem. Specifically, we render depth maps of the incomplete shape from a fixed set of viewpoints, and perform depth map completion in each view. Different from image-to-image translation networks that process each view separately, our novel multi-view completion net (MVCN) leverages information from all views of a 3D shape to help the completion of each single view. This enables MVCN to leverage more information from different depth views to achieve high accuracy in single depth view completion, and improve the consistency among the completed depth images in different views. Benefiting from the multi-view representation and novel network structure, MVCN significantly improves the accuracy of 3D shape completion in large-scale benchmarks compared to the state of the art.
\end{abstract}

\section{Introduction}

Shape completion is an important challenge in 3D shape analysis, serving as a building block in applications such as 3D scanning in robotics, autonomous driving, or 3D modeling and fabrication. While learning-based methods that leverage large shape databases have achieved significant advances recently, choosing a suitable 3D representation for such tasks remains a difficult problem. On the one hand, volumetric approaches such as binary voxel grids or distance functions have the advantage that convolutional neural networks can readily be applied, but including a third dimension increases the memory requirements and limits the resolutions that can be handled. On the other hand, point-based techniques provide a more parsimonious shape representation, and recently there has been much progress in generalizing convolutional networks to such irregularly sampled data. However, most generative techniques for 3D point clouds involve fully connected layers that limit the number of points and level of shape detail that can be obtained~\cite{fc_Achlioptas2018LearningRA,ref_cd,folding}.

In this paper, we propose to use a shape representation that is based on multi-view depth maps for shape completion. The representation consists of a fixed number of depth images taken from a set of pre-determined viewpoints. Each pixel is a 3D point, and the union of points over all depth images yields the 3D point cloud of a shape. This has the advantage that we can use several recent advances in neural networks that operate on images, like U-Net~\cite{unet} and 2D convolutional networks. In addition,  the number of points is not fixed and the point density can easily be increased by using higher resolution depth images, or more viewpoints.

Here we leverage this representation for shape completion. Our key idea to perform shape completion is to render multiple depth images of an incomplete shape from a set of pre-defined viewpoints, and then to complete each depth map using image-to-image translation networks. To improve the completion accuracy, we further propose a novel multi-view completion net (MVCN) that leverages information from all depth views of a 3D shape to achieve high accuracy for single depth view completion. In summary, our contributions are as follows:

\begin{itemize}
	\item A strategy to address shape completion by re-rendering multi-view depth maps to represent the incomplete shape, and performing image translation of these rendered views.
	\item A multi-view completion architecture that leverages information from all rendered views and outperforms separate depth image completion for each view. 
	\item More accurate 3D shape completion results than previous state of the art methods.
\end{itemize}			
\begin{figure*}[ht]
	\vspace{-0.1in}
	\begin{center}
		\includegraphics[width=\linewidth]{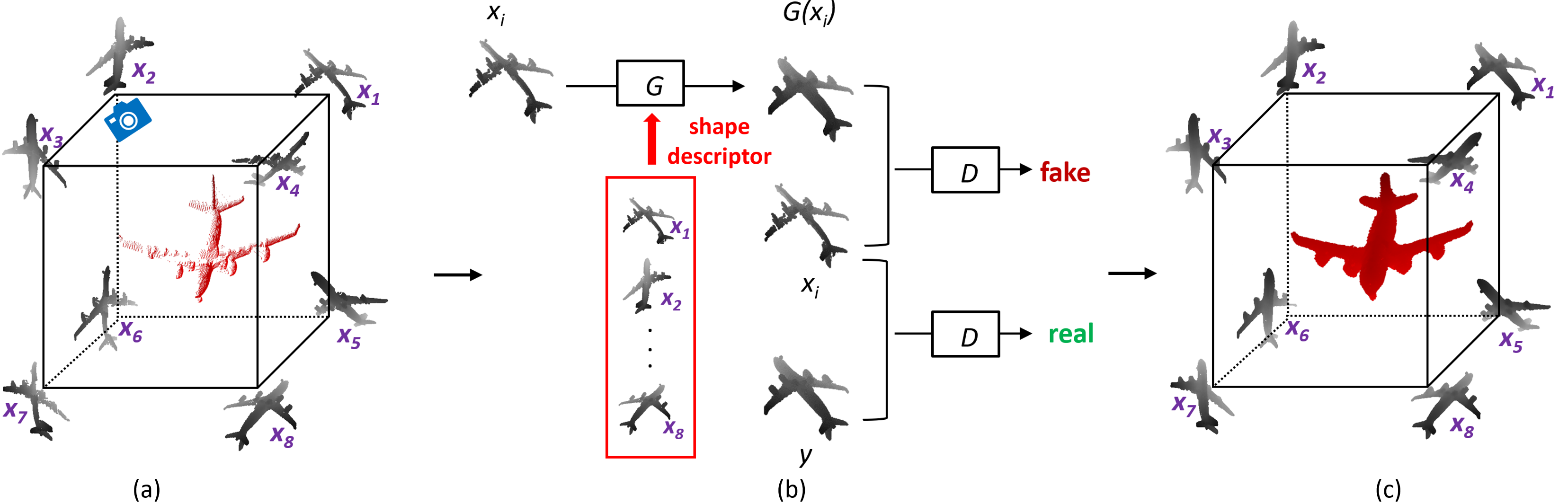}
	\end{center}
	\vspace{-0.12in}
	\caption{Overview of our approach. (a) We render 8 depth maps of an incomplete shape (shown in red) from 8 viewpoints on the corners of a cube; (b) These rendered 8 depth maps are passed through a multi-view completion net including an adversarial loss, which generates 8 completed depth maps; (c) We back-project the 8 depth maps into a completed 3D model.}
	\vspace{-0.12in}
	\label{fig:overview}
\end{figure*}	

\section{Related Work}

\noindent\textbf{Deep learning on 3D shapes.} Pioneering work on deep learning for 3D shapes has relied on volumetric representations~\cite{Maturana:2015:VXN,3D_ShapeNets}, which allow the straightforward application of convolutional neural networks. To avoid the computation and memory costs of 3D convolutions and 3D voxel grids, multi-view convolutional neural networks have also been proposed early for shape analysis~\cite{Qi_2016_VMV,su15mvcnn} such as recognition and classification. But these techniques cannot address shape completion. In addition to volumetric and multi-view representations, point clouds have also been popular for deep learning on 3D shapes. Groundbreaking work in this area includes PointNet and its extension~\cite{pointnet,pointnetpp}.

\noindent\textbf{3D shape completion.} Shape completion can be performed using volumetric grids, as proposed by Dai et al.~\cite{epn3d} and Han et al.~\cite{high_reso}, which are convenient for CNNs, like 3D-Encoder-Predictor CNNs for~\cite{epn3d} and encoder-decoder CNN for patch-level geometry refinement in~\cite{high_reso}. However, when represented with volumetric grids, data size grows cubically as the size of the space increases, which severely limits resolution and application. To address this problem, point based shape completion methods were presented, like \cite{fc_Achlioptas2018LearningRA, folding, ref_pcn}. The point completion network (PCN)~\cite{ref_pcn} is the state-of-the-art approach that extends the PointNet architecture~\cite{pointnet} to provide an encoder, followed by a multi-stage decoder that uses both fully connected~\cite{fc_Achlioptas2018LearningRA} and folding layers~\cite{folding}. They show that their decoder leads to better results than using a fully connected~\cite{fc_Achlioptas2018LearningRA} or folding based~\cite{folding} decoder separately. However, for these voxels or points based shape completion methods, the numbers of input and output voxels or points are still fixed. For example, the input should be voxelized on a $32^3$ grid~\cite{high_reso} and the output point cloud size is 2048~\cite{folding}, however, which can lead to loss of detail in many scenarios.


\noindent\textbf{3D reconstruction from images.} The problem of 3D shape reconstruction from single RGB images shares similarities with 3D shape completion, but is arguably even harder. While a complete survey of these techniques is beyond the scope of this paper, our work shares some similarities with the approach by Lin et al.~\cite{lin2018learning}. They use a multi-view depth map representation for shape reconstruction from single RGB images using a differentiable renderer. In contrast to their technique, we address shape completion, and our approach allows us to solve the problem directly using image-to-image translation. Soltani et al.~\cite{Soltani2017Synthesizing3S} do shape synthesis and reconstruction from multi-view depth images, which are generated by a variational autoencoder~\cite{Kingma2014AutoEncodingVB}. However, they do not consider the relations between the multi-view depth images of the same model in their generative net.

\noindent\textbf{Image translation and completion.} A key advantage of our approach is that it allows us to leverage powerful image-to-image translation architectures to address the shape completion problem, including techniques based on generative adversarial networks (GAN)~\cite{GAN}, and U-Net structures~\cite{unet}. Based on conditional GANs, image-to-image translation networks can be applied on a variety of tasks~\cite{pix2pix2016}. Satoshi et al.~\cite{comp_japan} and Portenier et al.~\cite{Portenier2018FaceshopDS} propose to use conditional GANs for image completion or editing. However, each image is completed individually in their networks. We propose a network that can combine information from other related images to help the completion of one single image.

\section{Method}

\subsection{Multi-view Representation}
\label{input}

As discussed above, high resolution completion is difficult to achieve by existing methods that operate on voxels or point clouds due to memory limitations or fully connected network structures. In contrast, multi-view representations of 3D shapes \cite{Qi_2016_VMV,su15mvcnn,Zhizhong2018VIP, Zhizhong2019seq} can circumvent these obstacles to achieve high resolution and dense completion. As shown in Fig.~\ref{fig:overview} (a), given an incomplete point cloud, our method starts from rendering 8 depth maps for this point cloud. Specifically, the renderings are generated by placing 8 virtual cameras at the 8 vertices of a cube enclosing the shape, all pointing towards the centroid of the shape. We also render 8 depth maps from the ground truth point cloud, and then we use these image pairs to train our network.

With this multi-view representation, the shape completion problem can be formulated as image-to-image translation, i.e., translating an incomplete depth map to a corresponding complete depth map, for which we can take full advantage of several recent advances in net structures that operate successfully on images, like U-Net architectures and GANs. After the completion net shown in Fig.~\ref{fig:overview}(b), we get 8 completed depth maps in Fig.~\ref{fig:overview}(c), which can be back-projected into a completed point cloud.

\subsection{Multi-view Depth Maps Completion}

In the completion problem, we learn a mapping from an incomplete depth map $ x_i $ to a completed depth map $ G(x_i) $, where $x_i$ is rendered from a partial 3D shape $S$, $i\in[1,V]$. We render $V$ views for each shape and expect to complete each depth map $x_i$ of $S$ as similar as possible to the corresponding depth map $y_i$ of the ground truth 3D shape $S_1$.

Although completing each of the $V$ depth maps of a 3D shape separately would be straightforward, there are two disadvantages. First, we cannot encourage consistency among the completed depth maps from the same 3D shape, which affects the accuracy of the resulting 3D shapes obtained by back-projecting the completed depth maps. Second, we cannot leverage information from other depth maps of the same 3D shape while completing one single depth map. This limits the accuracy of completing a single depth image, since views of the same 3D model share some common information that could be exploited, like global shape and local parts as seen from different viewpoints.

To resolve these issues, we propose a multi-view completion net (MVCN) architecture to complete one single depth image by jointly considering the global 3D shape information. In order to complete a depth image $x_i$ as similar as possible to the ground truth $y_i$ in terms of both low-frequency correctness and high-frequency structure, MVCN is designed based on a conditional GAN~\cite{GAN}, which is formed by an image-to-image net $G$ and a discriminator $D$. In addition, we introduce a shape descriptor $d$ for each 3D shape $S$ to contribute to the completion of each depth image $x_i$ from $S$, where $d$ holds global information of shape $S$. The shape descriptor $d$ is learned along with the other parameters in MVCN, and it is updated dynamically with the completion of all the depth images $x_i$ of shape $S$.

\subsection{MVCN Architecture}
\label{sec:net_structure}

\begin{figure}[htbp]
	\begin{center}
		\includegraphics[width=\linewidth]{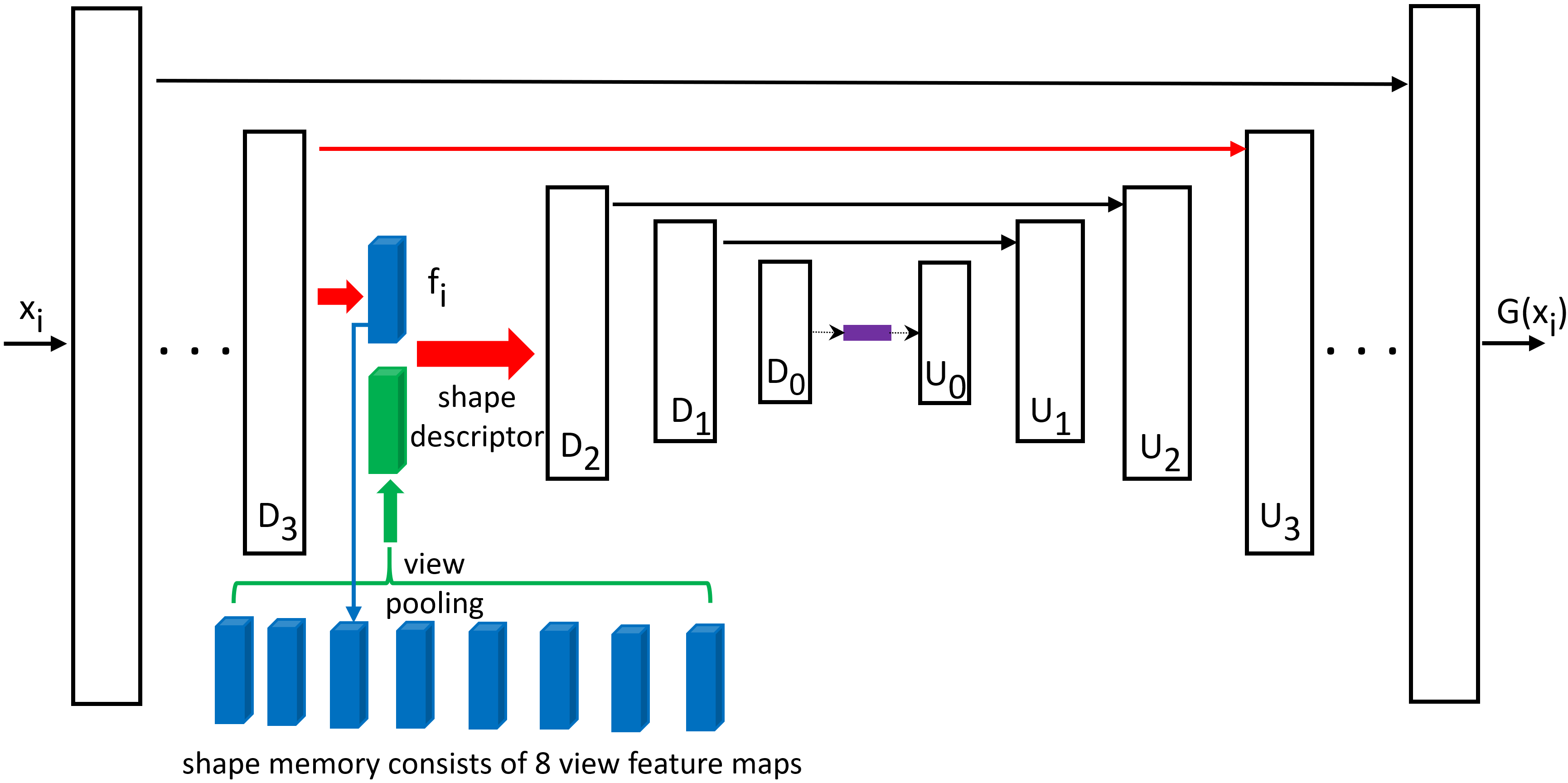}
	\end{center}
	\vspace{-0.12in}
	\caption{Architecture of MVCN. The shape descriptor represents the information of a 3D shape, which contributes to the completion of each single depth map from the 3D shape.}
	\vspace{-0.1in}
	\label{fig:mvcn}
\end{figure}

We use a U-Net based structure~\cite{unet} as our image-to-image net $G$, which has shown its effectiveness over encoder-decoder nets in many tasks including image-to-image translation~\cite{pix2pix2016}. 
Including our shape descriptor, we propose an end-to-end architecture as illustrated in Fig.~\ref{fig:mvcn}.

We adopt the generator and discriminator architecture of~\cite{pix2pix2016}. MCVN consists of 8 U-Net modules with an input resolution of $256\times 256$, and each U-Net module has two submodules, DOWN and UP. DOWN (e.g. $D_3$) consists of the Convolution-BatchNorm-ReLU layers~\cite{Ioffe2015BatchNA,Nair2010RectifiedLU}, and UP (e.g. $U_3$) consists of  UpReLU-UpConv-UpNorm layers. More details can be found in~\cite{pix2pix2016}.

In MVCN, DOWN modules are used to extract a view feature $f_i$ of each depth image $x_i$. For each 3D shape $S$, we learn a shape descriptor $d$ by aggregating all $V$ view features $f_i$ through a view-pooling layer. Since not all the features are necessary to represent the global shape, we use max pooling to extract the maximum activation in each dimension of all $f_i$ to form the shape descriptor, as illustrated in Fig.~\ref{fig:mvcn}. In addition, the shape descriptor $d$ is applied to contribute to the completion of each depth image $x_i$.

Specifically, for an input $x_i$ we employ the output of DOWN module $D_3$ as the view feature $f_i$, and insert the view-pooling layer after $D_3$. For each shape $S$ we use a shape memory to store all its $V$ view features $f_i$ as shown in Fig.~\ref{fig:mvcn}.  When we get $f_i$, we first use it to update the corresponding feature map in shape memory. For example, if $i=3$, the third feature map in shape memory will be replaced with $f_3$. Then we obtain the shape descriptor of $S$ in the current iteration by a view-pooling layer (max pooling all feature maps in the shape memory of $S$). This strategy dynamically keeps the best view features in all training iterations, as illustrated in Fig.~\ref{fig:mvcn}. Subsequently, we use shape descriptor $d$ to contribute to the completion of depth map $x_i$ by concatenating $d$ with view feature $f_i$ as the input of module $D_2$. This concatenated feature is also forwarded to module $U_3$ via a skip connection.

\subsection{Loss Function}
The objective of our conditional GAN is similar to image-to-image translation~\cite{pix2pix2016},
\begin{equation}
\begin{split}
\mathcal{L}_{cGAN}(G,D)= & \mathbb{E}_{x,y}[\log D(x,y)] + \\
& \mathbb{E}_{x}[\log(1- D(x,G(x))].
\end{split}
\label{eq_cgan}
\end{equation}

In our completion problem, we expect the completion net (G) could not only deceive the discriminator but also produce a completion result near the ground truth. Hence we combine the GAN objective with a traditional pixel-wise loss, such as L1 or L2 distance, which is consistent with previous approaches~\cite{pix2pix2016,pathakCVPR16context}. Since L1 is less prone to blurring than L2, and considering Eq.~\ref{eq_transf}, there is a linear mapping from a pixel in a depth image to a 3D point, we want to push the generated image to be near the ground truth in L1 sense rather than L2. Therefore, the loss of the completion net is
\begin{equation}
\mathcal{L}_{L1}(G)= \mathbb{E}_{x,y}[\|y-G(x)\|_1].
\label{eq_l1}
\end{equation}

Our final object in training is then
\begin{equation}
G^{*}= arg \mathop{min}\limits_{G} \mathop{max}\limits_{D}\mathcal{L}_{cGAN}(G,D) + \lambda \mathcal{L}_{L1}(G),
\label{eq_l1_gan}
\end{equation}

\noindent where $\lambda$ is a balance parameter that controls the contributions of the two terms.

\subsection{Optimization and Inference}

Unlike some approaches that focus on image generation~\cite{gene_more}, our method does not generate images from noise, which also makes our training stable, as mentioned in~\cite{comp_japan}. Similar to pix2pix~\cite{pix2pix2016}, we only provide noise in the form of dropout in our network.

To optimize our net, we follow the standard approach~\cite{GAN, pix2pix2016}.
The training of $D$ and $G$ is alternated, one gradient descent step on $D$, then one step on $G$. Minibatch SGD and the Adam solver~\cite{adam} are applied, with a learning rate of 0.0006 for $G$ and 0.000006 for $D$, which slows down the rate at which $D$ learns relative to $G$. Momentum parameters are $\beta_1 = 0.5, \beta_2 = 0.999$, and the batch size is 32.

During inference, we first run MVCN with all the 8 rendered views of an incomplete 3D shape to build the shape memory and extract the shape descriptor. Then we run the net again for the second time to complete each view leveraging the learned shape descriptor.

Our final target is 3D shape completion. Given a generated depth image $G(x_i)$, for each pixel $p$ at location $(x_p,y_p)$ with depth value $d_p$, we can back-project $p$ to a 3D point $P$ through an inverse perspective transformation,
\begin{equation}
P=R^{-1}(K^{-1}[x_p,y_p,d_p]^T-t),
\label{eq_transf}
\end{equation}
where $K$, $R$, and $t$ are the camera intrinsic matrix, rotation matrix, and translation vector respectively. Note that $K$, $R$, and $t$ are always known since these are the parameters of the 8 virtual cameras placed on the corners of a cube. The final shape is the union of the completed, back-projected point clouds from all 8 virtual views.

\section{Experiments}

In this section, we first describe the creation of a multi-category dataset to train our model, and then we illustrate the effectiveness of our method and the improvement of MCVN over a single view completion net (VCN) used as a baseline, where each view is completed individually without shape descriptor. Finally, we analyze the performance of our method, and make comparisons with existing methods. By default, we conduct the training of MVCN under the MVCN-Airplane600 (trained with the first 600 shapes of airplane in ShapeNet~\cite{shapenet}), and test it under the same 150 models involved in~\cite{ref_pcn}).

\subsection{Data Generation and Evaluation Metrics}
\label{data_generation}
We use synthetic CAD models from ShapeNet to create a dataset to train our model. Specifically, we take models from 8 categories: airplane, cabinet, car, chair, lamp, sofa, table, and vessel. Our inputs are partial point clouds. For each model, we extract one partial point cloud by back-projecting a 2.5D depth map (from a random viewpoint) into 3D, and render this partial point cloud into $V=8$ depth maps of resolution $256 \times 256$ as training samples. The reason why we use back-projected depth maps as partial point clouds instead of subsets of the complete point cloud is that our training samples are closer to real-world sensor data in this way. In addition, similar to other works, we choose to use a synthetic dataset to generate training data because it contains detailed 3D shapes, which are not available in real-world datasets. In the same way, we also render $V=8$ depth maps from the ground truth point clouds as the ground truth depth maps. More details of rendering and back-projecting depth maps are in the supplementary.

Similar to~\cite{ref_pcn}, here we also use the symmetric version of Chamfer Distance (CD)~\cite{ref_cd} to calculate the average closest point distance between the ground truth shape and the completed shape.

\subsection{Analysis of the Objective Function}

\begin{figure}[htbp]
	\begin{center}
		\includegraphics[width=\linewidth]{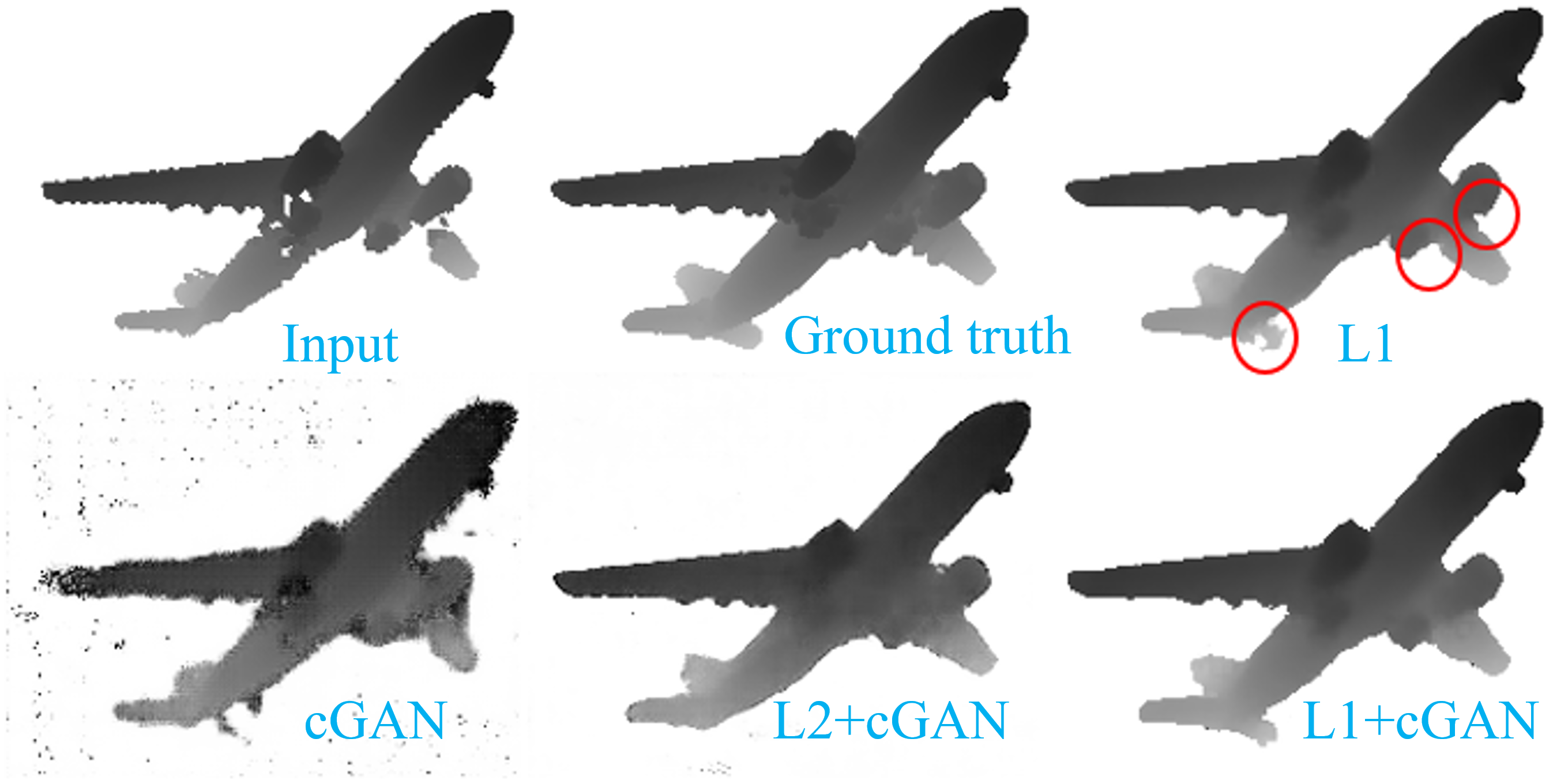}
	\end{center}
	\vspace{-0.12in}
	\caption{Completion results for different losses.}
	\vspace{-0.05in}
	\label{fig:loss}
\end{figure}

\begin{figure*}[t]
	\begin{center}
		\includegraphics[width=\linewidth]{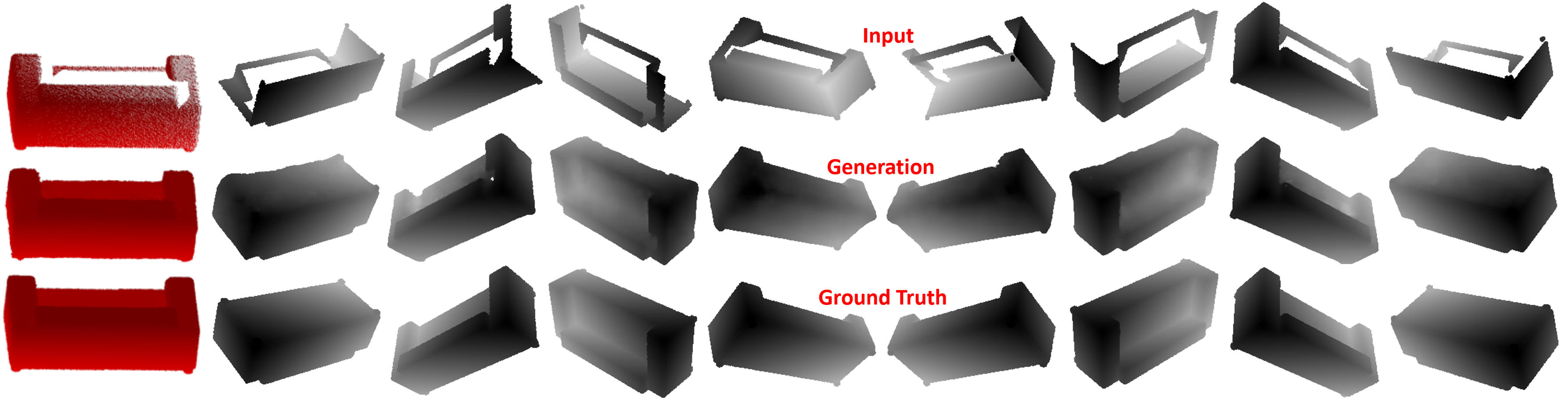}
	\end{center}
	\vspace{-0.12in}
	\caption{An example of the completion of sofa. The 1st row: incomplete point cloud and 8 depth maps of it;
		The 2nd row: generated point cloud and related 8 depth maps;
		The 3rd row: ground truth point cloud and its 8 depth maps.
	}
	\vspace{-0.12in}
	\label{fig:sofa_pip}
\end{figure*}

We conduct ablation studies to justify the effectiveness of our objective function for the completion problem. Table ~\ref{tab:loss_func}(a) shows the quantitative effects of these variations, and Fig.~\ref{fig:loss} shows the qualitative effects. The cGAN alone (bottom left, setting $\lambda= 0$  in Eq.~(\ref{eq_l1_gan})) gives very noisy results. L2+cGAN (bottom middle) leads to reasonable but blurry results. L1 alone (top right) also produces reasonable results, but we can find some visual defects, like some unfilled holes as marked, which makes the final CD distance higher than that of L1+cGAN. These visual defects can be reduced when including both L1 and cGAN in the loss function (bottom right). As shown by the example in Fig.~\ref{fig:sofa_pip}, the combination of L1 and cGAN can complete the depth images with high accuracy. We further explore the importance of the two components of the objective function for point cloud completion by using different weights ($ \lambda $ in Eq.~(\ref{eq_l1_gan})) of the L1 loss. In Table~\ref{tab:loss_func}(b), the best completion result is achieved when $\lambda=1$. We set $\lambda=1$ in our experiments.
\begin{table}[!htb]
	\begin{subtable}{.5\linewidth}
		\centering
		\begin{tabular}{|l|c|}
			\hline
			Loss &  Avg CD \\
			\hline\hline
			cGAN &	0.010729 \\
			L1 &		0.005672 \\
			L2 + cGAN &	0.006467 \\
			L1 + cGAN	& \textbf{0.005512} \\
			\hline
		\end{tabular}
		\caption{}
	\end{subtable}%
	\begin{subtable}{.5\linewidth}
		\centering
		\begin{tabular}{|l|c|}
			\hline
			$\lambda$ in Eq.~(\ref{eq_l1_gan}) & Avg CD \\
			\hline\hline
			$\lambda =50$ & 0.005748 \\
			$\lambda =10$ & 0.005665 \\
			$\lambda =1$	& \textbf{0.005512} \\
			$\lambda =0.5$ & 0.005541 \\
			\hline
		\end{tabular}
		\caption{}
	\end{subtable} 
	\vspace{-0.12in}
	\caption{Analysis of the objective function: average CD for different losses (a), and different $\lambda$ (b).}
	\vspace{-0.1in}
	\label{tab:loss_func}
\end{table}
\subsection{Analysis of the View-pooling Layer}
\noindent\textbf{Pooling methods.} We also study different view-pooling methods to construct the shape descriptor, including element-wise max-pooling and mean-pooling. According to our experiments, mean-pooling is not as effective as max-pooling to extract the shape descriptor for image completion, which is similar to the recognition problem~\cite{su15mvcnn}. The average CD is 0.005926 for mean-pooling, but that of max-pooling is 0.005512, so max-pooling is used.
\begin{table}[htbp]
	\begin{center}
		\begin{tabular}{|l|c|c|}
			\hline
			Position & Avg L1 distance & Avg CD \\
			\hline\hline
			$D_2$ 	 & 	\textbf{3.376642} &		\textbf{0.005512} \\
			$D_1$	 &	3.433185 &		0.005604 \\
			$D_0$	 &	3.500945 &		0.005919 \\
			Code &		3.477186 &		0.005836 \\
			\hline
		\end{tabular}
	\end{center}
	\vspace{-0.12in}
	\caption{Completion results for different positions of view-pooling layer}
	\vspace{-0.1in}
	\label{tab:position_sm}
\end{table}

\noindent\textbf{Position of the view-pooling layer.}  Here we insert the view-pooling layer into different positions to extract the shape descriptor and further evaluate its effectiveness, including $D_2$, $D_1$, and $D_0$, which are marked in Fig.~\ref{fig:mvcn}. Intuitively, the shape descriptor would have the biggest impact on the original network if we place the view-pooling layer before $D_2$, and the experimental results illustrate this in Table~\ref{tab:position_sm}, where both average L1 distance and CD are the lowest. We also try to do view pooling after $D_0$ and concatenate the shape descriptor with the latent code (marked in purple in Fig.~\ref{fig:mvcn}) and then pass them through a fully connected layer, but experiments show that the shape descriptor will be ignored since both the average L1 distance and CD do not decrease compared with single view completion net (average L1 distance is 3.473643 and  CD is 0.005839 in Table~\ref{tab:5_cmp_vcn_mvcn}).
\begin{table}[htbp]$  $
	\begin{center}
		\begin{tabular}{|c|c|}
			\hline
			Model Name & Avg L1 Distance \\
			\hline\hline
			{MVCN-V3} &	3.794273 \\
			{MVCN-V8-3} & 3.616740 \\	
			\hline\hline
			{MVCN-V5} &	3.564278 \\
			{MVCN-V8-5} & 3.397558 \\
			\hline
		\end{tabular}
	\end{center}
	\vspace{-0.12in}
	\caption{Average L1 distance for different numbers of views in view-pooling.}
	\vspace{-0.1in}
	\label{tab:num_views}
\end{table}
\begin{figure}[htbp]
	\begin{center}
		\includegraphics[width=\linewidth]{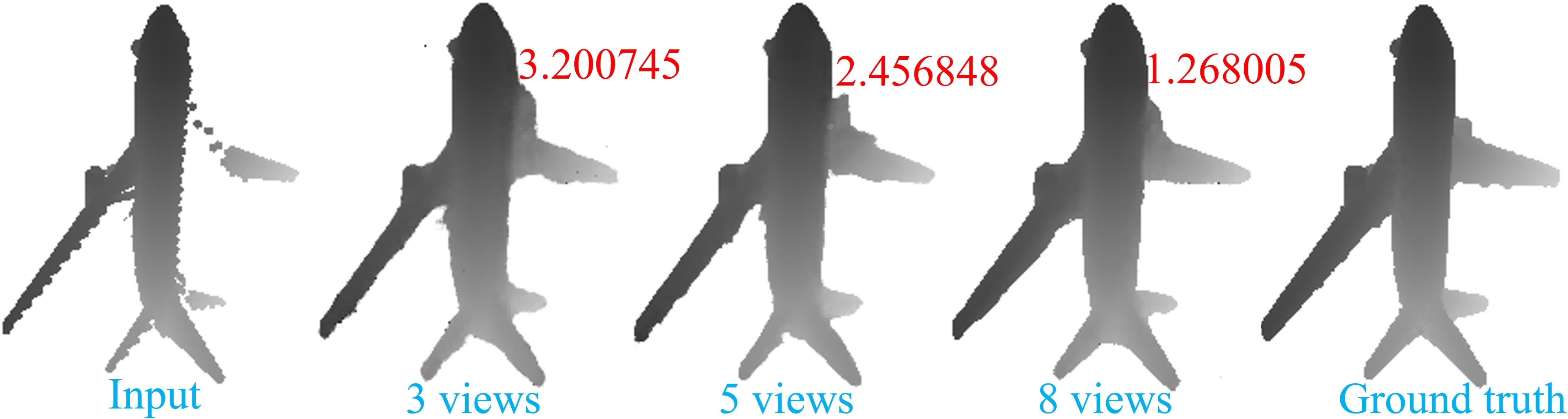}
	\end{center}
	\vspace{-0.12in}
	\caption{Completion results for different numbers of views in view-pooling.}
	\vspace{-0.1in}
	\label{fig:vis_num_view}
\end{figure}
\label{sect_number_view}

\noindent\textbf{Number of views in view-pooling.} We also analyze the effect of the number of views used in view-pooling. In Table~\ref{tab:num_views}, MVCN-V3 was trained with 3 depth images (No.1, 3, 5) of the 8 depth images of each 3D model, and MVCN-V5 was trained with 5 depth images (No. 1, 3, 5, 6, 8). MVCN-V8-3 and MVCN-V8-5 were trained with all the 8 depth images, but were tested with 3 views and 5 views respectively. In order to make fair comparisons, we took the 1st, 3rd, and 5th view images to test MVCN-V8-3 and MVCN-V3, and 1st, 3rd, 5th, 6th, 8th to test MVCN-V8-5 and MVCN-V5. The results show that the completion of one single view will be better when we increase the number of views, which means other views are helpful for the completion of one single view, and the more the views, the higher the completion accuracy. Fig.~\ref{fig:vis_num_view} shows an example of the completion. As we increase the number of views in view-pooling, the completion results are improved.%
\begin{figure*}[th]
	\begin{center}
		\includegraphics[width=\linewidth]{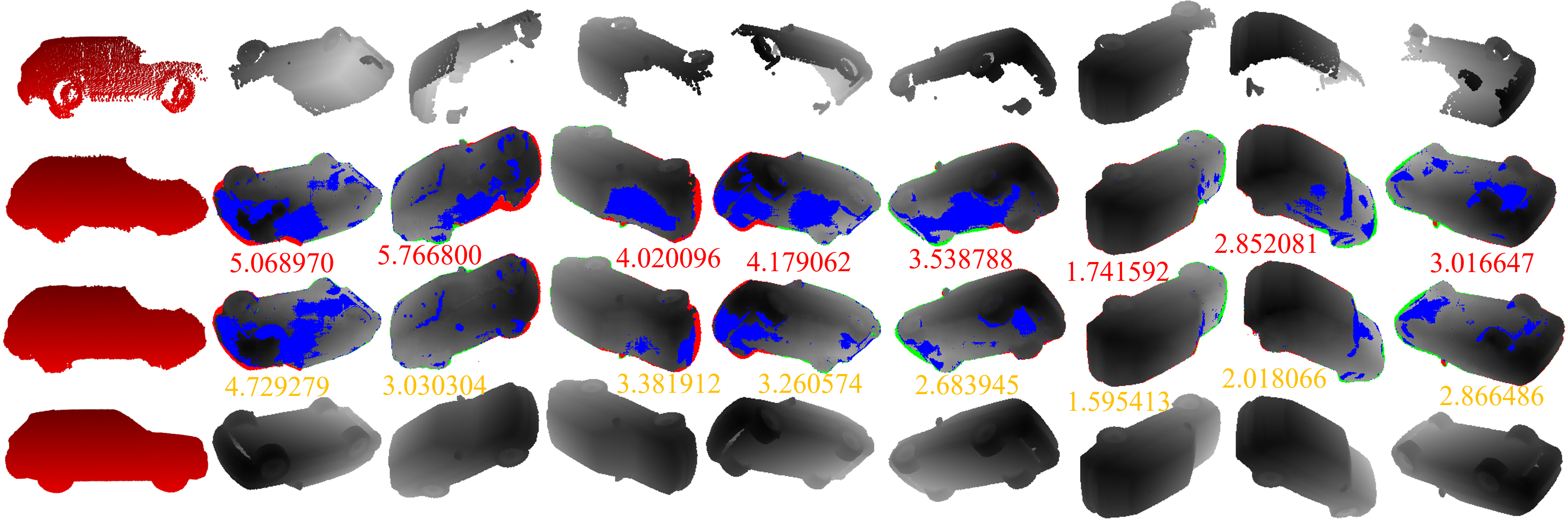}
	\end{center}
	\vspace{-0.2in}
	\caption{Visual comparison between VCN and MVCN. Starting from the partial point cloud in the first row, VCN and MVCN perform completions of depth maps in the second and third row, respectively, where the completed point clouds are also shown. We use colormaps (from blue to green to red) to highlight the pixels with bigger errors than 10 in terms of L1 distance. Ground truth data is in the last row. MVCN achieves lower L1 distance on all the 8 depth maps.
	}
	\vspace{-0.1in}
	\label{fig:mvcn_vcn}
\end{figure*}
\subsection{Improvements over Single View Completion}
\label{consist_relation}

\noindent\textbf{Pervasive improvements on L1 distance and CD}. From Table~\ref{tab:5_cmp_vcn_mvcn}, we find significant and pervasive improvements over single view completion net (VCN) on both average L1 distance and CD on different categories. Nets in Table~\ref{tab:5_cmp_vcn_mvcn} were trained with 600 3D models for airplane, 1600  for lamp, and 1000 for other categories. We use 150 models of each category to evaluate our network, the same test dataset in~\cite{ref_pcn}. We further conduct visual comparison with VCN in Fig.~\ref{fig:mvcn_vcn}, where we can see MVCN can achieve higher completion accuracy with the help of the shape descriptor.
\begin{table}[ht]
	\begin{center}
		\begin{tabular}{|l|c|c|}
			\hline
			Model &	Avg L1 Distance & Avg CD \\
			\hline\hline
			MVCN-Airplane600 & 	3.376642	 & 0.005512 \\
			MVCN-Airplane1200	 & 3.156200	 & 0.005273 \\
			MVCN-Lamp1000  & 	6.660511 & 	0.012012 \\
			MVCN-Lamp1600	 & 6.245297 & 	0.010576 \\
			VCN-Lamp1000	 & 6.763339 & 	0.012091 \\
			VCN-Lamp1600	 & 6.430318 & 	0.012007 \\
			\hline
		\end{tabular}
	\end{center}
	\vspace{-0.12in}
	\caption{Improvements while increasing training samples.}
	\vspace{-0.1in}
	\label{tab:training_samp}
\end{table}

\noindent\textbf{Better generalization capability}. Table~\ref{tab:training_samp} shows that we can improve the performance of VCN and MVCN while increasing the number of training samples. We find that the performance differences between MVCN-Lamp1000 and VCN-Lamp1000 are not obvious. The reason is that there are relatively large individual differences among lamp models in ShapeNet, and the completion results are bad in several unusual lamp models in the test set. For these models, the comparisons between VCN and MVCN are less meaningful, so the improvement is not obvious. But this can be solved when we add another 600 training samples in training. MVCN-Lamp1600 has a bigger improvement than VCN-Lamp1600 on average L1 distance and CD, which indicates a better generalization capability of MVCN. 
\begin{table*}[t]
	\begin{tabular}{c|c|c|c|c|c|c|c|c|c}
		{Model} & \multicolumn{9}{c}{Average L1 Distance}\\                    
		\hline
		& Avg & Airplane & Cabinet  & Car      & Chair    & Lamp     & Sofa     & Table    & Vessel   \\
		VCN                    &  5.431036   & 3.473643 & 4.304635 & 3.858853 & 7.644824 & 6.430318 & 5.716992 & 7.572865 & 4.44616  \\
		MVCN                   &  \textbf{5.102478}   & \textbf{3.376642} & \textbf{3.991407} & \textbf{3.609639} & \textbf{7.143200}   & \textbf{6.245297} & \textbf{5.284686} & \textbf{7.155616} & \textbf{4.013339} \\
		\hline\hline
	\end{tabular}
	\begin{tabular}{c|c|c|c|c|c|c|c|c|c}
		{Model} & \multicolumn{9}{c}{\begin{tabular}[c]{@{}c@{}}Mean Chamfer Distance per point\end{tabular}}                                          \\
		\hline
		& Avg & Airplane & Cabinet  & Car      & Chair    & Lamp                                                  & Sofa     & Table    & Vessel   \\
		VCN                    &  0.008800   & 0.005839 & 0.007297 & 0.006589 & 0.010398 & 0.012007                                              & 0.009565 & 0.009371 & 0.009334 \\
		MVCN                   &  \textbf{0.008328 }  & \textbf{0.005512} & \textbf{0.007154} & \textbf{0.006322} & \textbf{0.010077} & \begin{tabular}[c]{@{}c@{}}\textbf{0.010576}\end{tabular} & \textbf{0.009174} & \textbf{0.009020} & \textbf{0.008790} \\
		\hline
	\end{tabular}
	\caption{Comparison of average L1 Distance and mean Chamfer Distance between VCN and MCVN.}
	\vspace{-0.1in}
	\label{tab:5_cmp_vcn_mvcn}
\end{table*}

\subsection{Comparisons with the State-of-the-art} 
\label{sect_cmp_other}

\noindent\textbf{Baselines}. Some previous completion methods need prior knowledge of the shape~\cite{semantics}, or assume more complete inputs~\cite{Poisson}, so they are not directly comparable to our method. Here we compare MVCN with several strong baselines. \textbf{PCN-CD}~\cite{ref_pcn} trained with point completion net with CD as loss function, is the state of the art while this work was developed. \textbf{PCN-EMD} uses Earth Mover’s Distance (EMD)~\cite{ref_cd} as loss function, but it is intractable for dense completion due to the calculation complexity of EMD. The encoders of \textbf{FC}~\cite{fc_Achlioptas2018LearningRA}, \textbf{Folding}~\cite{folding} are the same with PCN-CD, but decoders are different, a 3-layer fully-connected network for FC, and folding-based layer for Folding. \textbf{PN2} uses the same decoder, but the encoder is PointNet++~\cite{pointnetpp}. \textbf{3D-EPN}~\cite{epn3d} is a representative of the class of volumetric completion methods.  For fair comparison, the distance field outputs of 3D-EPN are converted into point clouds as mentioned in~\cite{ref_pcn}. TopNet \cite{topnet} is a recent point-based method, but it can only generate sparse point clouds because their decoder mostly consists of multilayer perceptron networks, which limits the number of points they can process.
\begin{table*}[h]
	\begin{center}
		\begin{tabular}{c|c|c|c|c|c|c|c|c|c}
			{Model} & \multicolumn{9}{c}{Mean Chamfer Distance per point}                                          \\
			\hline
			& Avg & Airplane & Cabinet  & Car      & Chair    & Lamp                                                  & Sofa     & Table    & Vessel   \\
			
			3D-EPN                                         & 2.0147                     & 1.3161                     & 2.1803                     & 2.0306                     & 1.8813                     & 2.5746                     & 2.1089                     & 2.1716                     & 1.8543                     \\
			FC                                         & 0.9799                     & 0.5698                     & 1.1023                     & 0.8775                     & 1.0969                     & 1.1131                     & 1.1756                     & 0.9320                     & 0.9720                     \\
			Folding                                    & 1.0074                     & 0.5965                     & 1.0831                     & 0.9272                     & 1.1245                     & 1.2172                     & 1.1630                     & 0.9453                     & 1.0027                     \\
			PN2                                        & 1.3999                     & 1.0300                     & 1.4735                     & 1.2187                     & 1.5775                     & 1.7615                     & 1.6183                     & 1.1676                     & 1.3521                     \\
			PCN-CD                                     & 0.9636                     & 0.5502                     & 1.0625                     & 0.8696                     & 1.0998                     & 1.1339                     & 1.1676                     & \textbf{0.8590 }                    & 0.9665                     \\
			PCN-EMD                                    & 1.0021                     & 0.5849                     & 1.0685                     & 0.9080                     & 1.1580                     & 1.1961                     & 1.2206                     & 0.9014                     & 0.9789                     \\
			MVCN                                       &           \textbf{0.8298}                   &\textbf{ 0.5273  }                   &\textbf{ 0.7154  }                   & \textbf{0.6322}                     & \textbf{1.0077 }                    & \textbf{1.0576  }                  & \textbf{0.9174 
			}                   & 0.9020                     &\textbf{ 0.8790 }                   
		\end{tabular}
		\vspace{-0.1in}
		\caption{Comparison with the state-of-the-art in terms of mean CD (multiplied by 100) per point over multiple categories.}
		\vspace{-0.1in}
		\label{tab:cmp_other}
	\end{center}
\end{table*}
\begin{figure}[htbp]
	\begin{center}
		\includegraphics[width=0.8\linewidth]{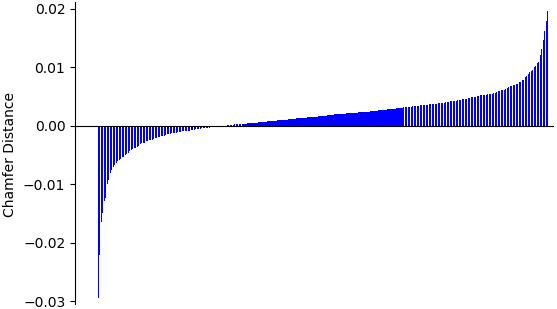}
	\end{center}
	\vspace{-0.1in}
	\caption{Comparison between MVCN and PCN-CD.}
	\vspace{-0.1in}
	\label{fig:cmp_objects}
\end{figure}

\begin{figure*}[h]
	\begin{center}
		\includegraphics[width=0.923\linewidth]{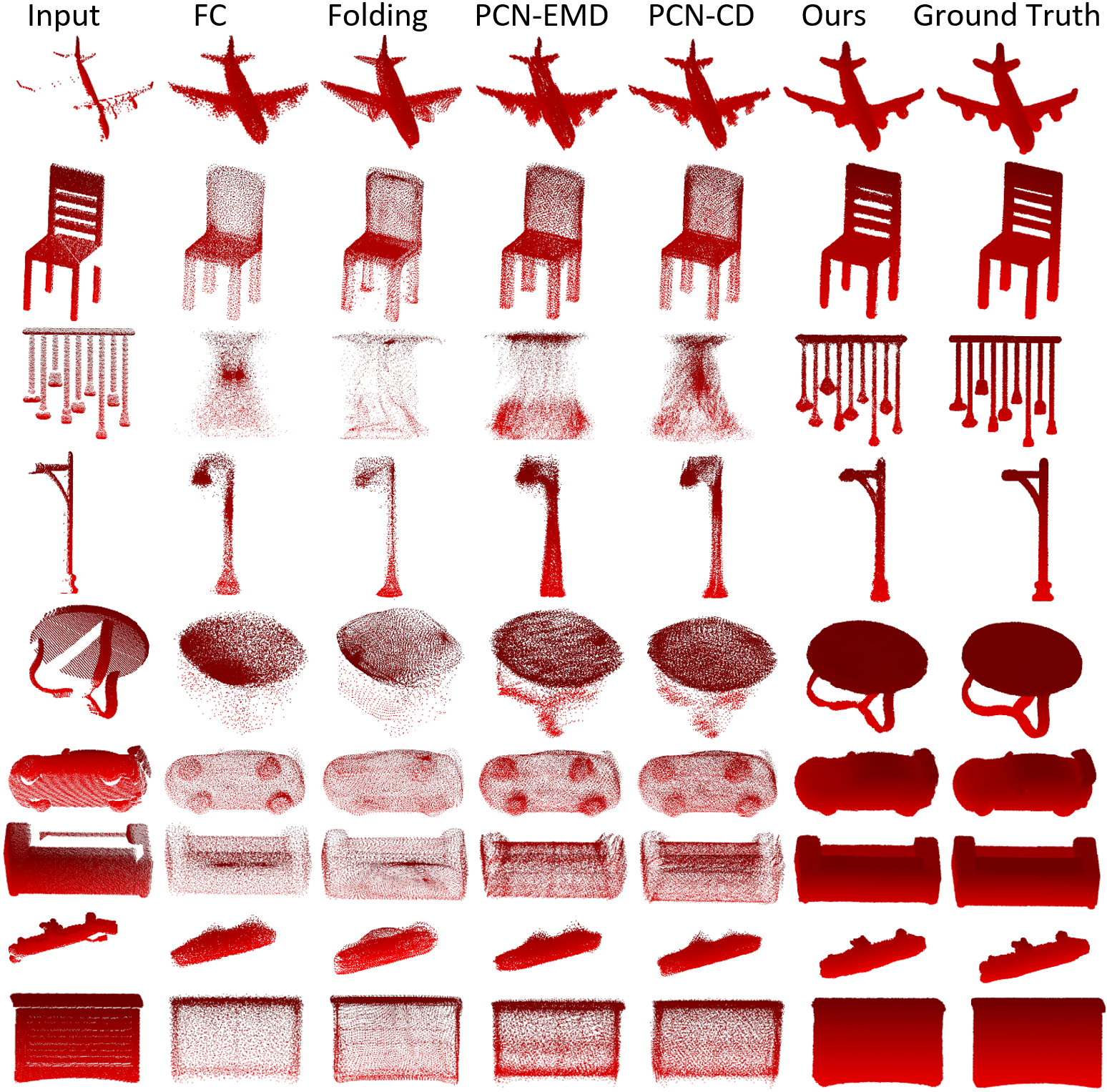}
	\end{center}
	\caption{Qualitative completion on ShapeNet, where MVCN can complete complex shapes with high resolution.}
	\label{fig:vis_cmp}
\end{figure*}

\noindent\textbf{Comparisons}. As done in~\cite{ref_pcn}, we use the symmetric version of CD to calculate the average closest point distance, where ground truth point clouds and generated point clouds are not required to be the same size, which is different from EMD ~\cite{ref_cd}. For point-based methods like PCN \cite{ref_pcn}, the input is sampled and the output size is fixed, which makes the calculation of EMD relatively easy. Different from these methods, the number of output points of our approach is not fixed, which would require resampling our output to compute the EMD. CD is more suitable for a fair comparison among different techniques. Table~\ref{tab:cmp_other} shows the quantitative results, where the completion results of other methods are from~\cite{ref_pcn}. Our method achieves the lowest CD across almost all object categories. A more detailed comparison with PCN-CD is in Fig.~\ref{fig:cmp_objects}, where the height of the blue bar indicates the amounts of improvement of our method over PCN-CD on each object. Our model outperforms PCN on most objects. Fig.~\ref{fig:vis_cmp} shows the qualitative results. Our completions are denser, and we recover more details in the results. Another obvious advantage is that our method can complete shapes with complex geometry, like the 2nd to 4th objects, but other methods fail to recover these shapes.

\subsection{Completion Results on KITTI}
Our goal is to obtain high quality and high resolution shape completion from data similar to individual range scans focused on individual objects. Hence we obtain incomplete data using synthetic depth images, which is similar to data from RGB-D cameras. However, for data like KITTI, which is extremely sparse and does not contain ground truth, the usual objective is to obtain rough not high resolution completion. Our method performs reasonably well on KITTI data, as shown in Fig.~\ref{fig:kitti}. More completion results on noisy, sparse, and occluded data can be found in the supplementary material.
\begin{figure}[h]
	\begin{center}
		\includegraphics[width=\linewidth]{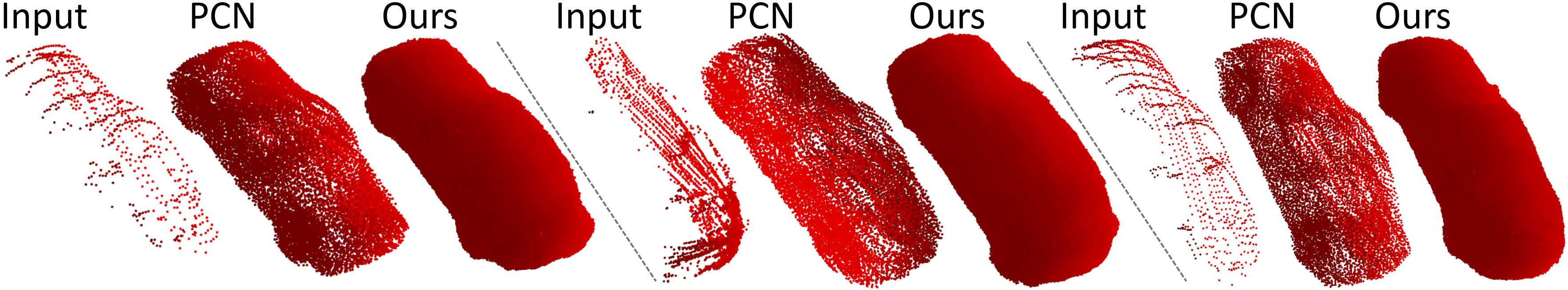}
	\end{center}
	\vspace{-0.1in}
	\caption{Completion results on KITTI.}
	\vspace{-0.1in}
	\label{fig:kitti}
\end{figure}
\section{Conclusion}
We have presented a method for shape completion by rendering multi-view depth maps of incomplete shapes, and then performing image completion of these rendered views. Our multi-view completion net shows significant improvements over a baseline single view completion net across multiple object categories. Experiments show that our view based representation and novel network structure can achieve better results with less training samples, perform better on objects with complex geometry, and generate higher resolution results than previous methods.

{\small
	\bibliographystyle{ieee}
	\bibliography{egbib}
}

\newpage

\pagebreak

\appendix
\section*{Appendices}
\addcontentsline{toc}{section}{Appendices}
\renewcommand{\thesubsection}{\Alph{subsection}}

In this document, we provide additional experimental results and technical details of the main paper.

\subsection{Completion Results on Noisy, Sparse and Occluded Point Clouds}
\begin{figure}[h]
	\begin{center}
		\includegraphics[width=\linewidth]{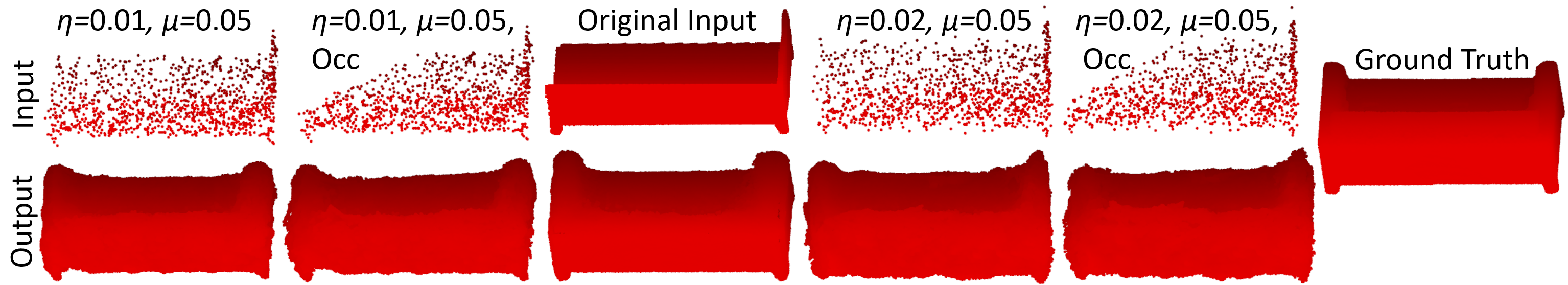}
	\end{center}
	\vspace{-0.1in}
	\caption{Completion results on noisy, sparse and occluded inputs.}
	\vspace{-0.05in}
	\label{fig:cmp}
\end{figure}

Since there is no ground truth on KITTI, we also conduct experiments to evaluate the performance of our method on noisy, sparse and occluded inputs in Fig.~\ref{fig:cmp}. For the ground truth point cloud (sofa), we render a depth image from a random viewpoint, whose back-projection is labeled `Original Input', and then perturb the depth map with Gaussian noise whose standard deviation is $\eta$ times the scale of the depth measurements. We then randomly sample the input point clouds with a factor $\mu$. Besides self-occlusion, we also consider the target may be occluded by other objects in the wild, and in Fig.~\ref{fig:cmp}, `Occ' in the 2nd and the 5th column means that we further remove 10$\%$ of the input points. Note that our model is not trained with these noisy, sparse, and occluded examples, but it is still robust to them. 

\subsection{Analysis of the Number of Views in View-pooling}

\begin{figure}[htbp]
	\begin{center}
		\includegraphics[width=\linewidth]{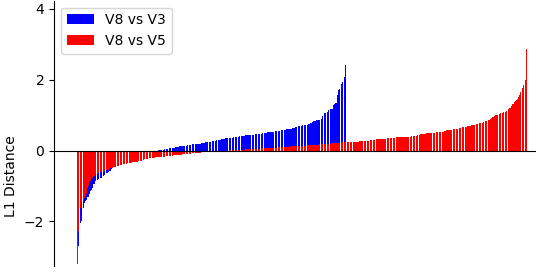}
	\end{center}
	\caption{Improvements of 8 views over 3 and 5 views in view-pooling.}
	\label{fig:diag_num_view}
\end{figure}

More experimental results for Section~\ref{sect_number_view}. We further show the improvements on L1 distance for all view images of test dataset in Fig.~\ref{fig:diag_num_view}. The x-axis represents different view images. It should be mentioned that the same $x$ represents different view images for `V8 vs V3' and ‘V8 vs V5’, considering the test dataset has 150 3D models, so 450 view images are used to test ‘V8 vs V3’, and 750 view images are used to test ‘V8 vs V5’). The height of the blue bar indicates the amounts of improvement of 8 views over 3, and the red bar indicates the improvement of 8 views over 5. Positive values mean the L1 distance is lower while using 8 views. Since the training dataset is relatively small (600 3D models for training and 150 3D models for testing), our net performs bad on several unusual models in testing dataset, which fall on the boundary in Fig.~\ref{fig:diag_num_view}. Comparisons on boundary instances are not meaningful. Apart from these, for most view images, we decrease L1 distance by increasing the number of views in view-pooling. More views mean the shape descriptors are more helpful.

\subsection{Failure Cases}
\begin{figure}[h]
	\begin{center}
		\includegraphics[width=0.8\linewidth]{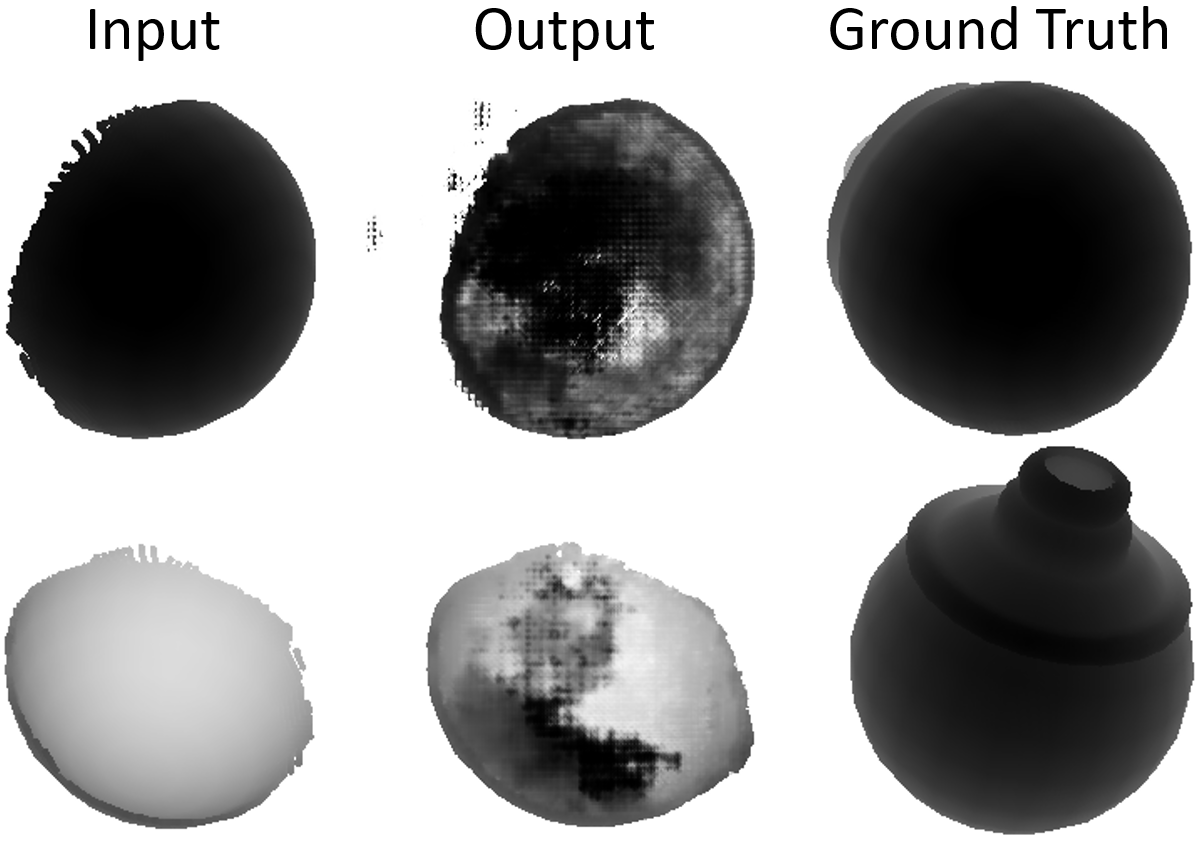}
	\end{center}
	\caption{Failed depth completions of lamps.}
	\label{fig:failure_case}
\end{figure}

While in general our methods perform well, we observe our models fail to complete several challenging input depth maps, which do not provide enough information for inference. For example, Fig.~\ref{fig:failure_case} shows two failed completions of lamps, where we cannot extract useful information from the depth inputs to infer the whole shape. These cases mostly occur in lamp objects due to complex geometry and large individual differences among lamp objects. The reconstruction of lamp is also the most challenging task, as mentioned in \cite{Mescheder2018OccupancyNL}.  

\subsection{Rendering and Back-projecting Depth Maps}

\textbf{Render multi-view depth maps.} First, for each 3D model, we move its center to the origin. Most models in modern online repositories, such as ShapeNet and the 3D Warehouse, satisfy this requirement that models are upright oriented along a consistent axis, and some previous completion or recognition methods also follow the same assumption~\cite{3D_ShapeNets,ref_pcn}.  With this assumption, the center consists of the midpoints along $x, y, z$ axis. Then, each model is uniformly scaled to fit into a consistent sphere (radius is 0.2) and the scale factor is the maximum length along $x, y, z$ axis divided by radius. Finally, we render 8 depth maps for each partial point cloud, as mentioned in Section~\ref{input}. In this way, all the shapes occur at the center of depth images. We also render 8 depth maps of the ground truth shape and use these image pairs to train our net. 

\textbf{Back-project multi-view depth maps into a point cloud}. We fuse the generated depth maps into a completed point cloud and apply voting algorithm to remove outliers. Specifically, we reproject each point of one view into the other 7 views, and if one point falls on the shape of other views, we add one vote for it. The initial vote number for each point is 1, and we set a vote threshold of 7 to decide whether this point is valid or not. Furthermore, radius outlier removal method is used to remove noisy points that have few neighbors (less than 6) in a given sphere (radius is 0.006) around them. We also back-project multi-view ground truth depth maps as ground truth point cloud, and restore the generated model to its original size to calculate CD.
\end{document}